\documentclass[letterpaper, 10 pt, conference]{ieeeconf}  

\IEEEoverridecommandlockouts                              

\usepackage{graphics} 
\usepackage{epsfig} 
\usepackage{mathptmx} 
\usepackage{times} 
\usepackage{amsmath} 
\usepackage{amssymb}  
\usepackage{textcomp}

\title{\LARGE \bf
Automatic Snake Gait Generation Using Model Predictive Control
}

\author{Emily Hannigan, Bing Song, Gagan Khandate, Maximilian Haas-Heger, Ji Yin and Matei Ciocarlie
\thanks{*This work was supported in part by ONR grant N00014-19-1-2062 and NSF CAREER award IIS-1551631. \newline \indent All authors are with the Dept. of Mechanical Engineering, Columbia University, NY. E-mail: {\tt\small \{ejh2192, bs2664, gk2496, mkh2149, jy2917, mtc2103\}@columbia.edu}}
}

\begin{document}

\maketitle
\thispagestyle{empty}
\pagestyle{empty}

\begin{abstract}

In this paper, we propose a method for generating undulatory gaits for snake robots. Instead of starting from a pre-defined movement pattern such as a serpenoid curve, we use a Model Predictive Control (MPC) approach to automatically generate effective locomotion gaits via trajectory optimization. An important advantage of this approach is that the resulting gaits are automatically adapted to the environment that is being modeled as part of the snake dynamics. To illustrate this, we use a novel model for anisotropic dry friction, along with existing models for viscous friction and fluid dynamic effects such as drag and added mass. For each of these models, gaits generated without any change in the method or its parameters are as efficient as Pareto-optimal serpenoid gaits tuned individually for each environment. Furthermore, the proposed method can also produce more complex or irregular gaits, e.g. for obstacle avoidance or executing sharp turns.

\end{abstract}

\section{INTRODUCTION}\label{sec:introduction}
Snake robots have great potential in challenging environments due to their many degrees of freedom (DoF), providing them with the ability to adapt to cluttered and complex terrain \cite{Hirose_1993, Pettersen_2017_review, Gong_2016}. However, the large number of DoF also makes generating efficient locomotion gaits a significant challenge.  As a result, significant work has focused on snake robot gait synthesis, starting with the mathematical approximation of their biological counterparts. 

The serpenoid curve \cite{Hirose_1993} and the serpentine curve \cite{Ma_1999_analysis} approximate snake lateral undulation for locomotion over flat surfaces. More recently, central pattern generators (CPGs) inspired by neural activity have been used to produce rhythmic patterns as snake gaits \cite{Ijspeer_2008_review}. Realization of these gaits on a robot mechanism is usually done through position  control of the robot's joint angles \cite{Hirose_1993, Prautsch_1999, Pettersen_2017_review, Whitman_2016}. Based on predefined gait patterns like the serpenoid curve, perception-driven obstacle-aided snake robots are able to perform complex gaits against their terrain like a biological snakes \cite{sanfilippo_2017}. 

We characterize different environments by what we refer to as different \textit{reaction forces}; these include dry friction (for land snakes), viscous friction (for undulatory movement in water at a regime with a low Reynolds number), and fluid dynamic effects like added mass (for movement in water at a regime with a high Reynolds number). Predefined movement patterns (such as a serpenoid curve) generally require hand-tuning in order to achieve efficient gaits in each of these environments \cite{saito2002, Kelasidi2018LocomotionEO}. Finding the serpenoid gait parameters that achieve a specific velocity, or that minimize power consumption for a given velocity, is a manual process. The resulting parameters rarely transfer well to different environmental conditions. Even compliant controllers that exhibit remarkable adaptability to obstacles \cite{Whitman_2016} are still designed or hand-tuned specifically for an underlying model of surface friction. Synthesising efficient gaits across different environmental conditions, without changing the locomotion algorithm or its parameters, remains a challenging problem. 

In this paper, we propose an optimal control approach to gait generation that does not start from a pre-defined movement pattern. Instead, based on a model of snake dynamics, the algorithm attempts to produce trajectories that minimize a given cost function. By choosing an appropriate cost function (e.g. distance to a goal), we show that this approach produces effective locomotion gaits. Importantly, this approach produces efficient gaits accross multiple different environments: as long as the appropriate environment reaction forces are modeled as part of the dynamics, the algorithm automatically produces an efficient gait adapted to the environment type. The resulting gaits are comparable to the baseline of the Pareto-optimal serpenoid gaits. Furthermore, our appoach readily transfers between different environments, with no tuning or change in parameters.

While matching the performance of a pre-defined serpenoid gait for straight-line locomotion, the optimal control approach can also synthesise more complex gaits: we show that this approach can generate gaits for passing through narrow tunnels, or for taking sharp turns. In each case, the only modification needed is in the cost function, which must reflect the requirements of each task, such as maintaining distance from the obstacles or reaching a desired point.

Overall, the main contributions of our paper include:
\begin{itemize}
\item We show that optimal control methods can generate effective undulatory gaits for snake robots without starting from a  predefined movement pattern. When this approach is applied using different models for environment reaction forces, it produces different gaits, each suited to its environment, without any change of the algorithm or tuning of its parameters. These gaits show both locomotion and energy efficiency.
\item We introduce a novel model of anisotropic Coulomb friction, and use it to model land snakes that exhibit side slip. Together with existing models for anisotropic viscous friction, drag force and added mass effects in fluids, we can model a wide range of environments for snake locomotion.
\item We also show that snake gaits obtained via optimal control exhibit additional desirable properties, such as the ability to perform sharp turns, as well as pass through narrow obstacles without making contact.
\end{itemize}

\section{Related Work}\label{sec:related_work}

Previous work on snake gait generation generally focuses on producing compliant motion across unstructured surface terrains or in fluids. To achieve such compliant motion, low level torque controllers \cite{Rollinson_2014} and shape-based
compliant control \cite{Whitman_2016} are used to adapt snake gaits through an irregularly spaced peg array. Integral line-of-sight guidance control \cite{Pettersen_2017_review} is applied to underwater snake robots to adapt to ocean currents of unknown direction and magnitude. Sliding mode control \cite{Mukherjee_2017}, mixed integer programming \cite{Kon_2016}, reinforcement learning \cite{snake_gaits_ppo} and CPGs \cite{Ijspeer_2008_review, cpg_obstacle} have also been investigated to produce gaits.

In adition to adapting to unstructured terrain, gaits must be efficient to be practical in real world applications. Gait efficiency has been improved by  \cite{Tesch_2011, Kelasidi_2018, snake_gaits_ppo}. Recently, Sartoretti et al. \cite{Sartoretti_2018} develped a decentralized control method that improved the locomotion efficiency of the compliant controller developed in \cite{Whitman_2016} by $40\%$.

In this paper we use an optimal control method, i.e., MPC, to automatically generate effective gaits across multiple environments without retuning of parameters. The MPC algorithm we use is iterative linear-quadratic-regulator (iLQR) \cite{tassa2012iLQR}. In work performed concurrently with ours \cite{nonhoff2019economic}, MPC was also used for snake gait generation. Although their work is similar to ours, we distinguish our work with the use of more accurate snake dynamics and experimentations with a wide range of accurate reaction forces. We also show our progress towards real time implementation.

Existing work on modeling undulatory locomotion either assumes a sideslip constraint \cite{Ostrowski_1998}, or anisotropic friction applied to the robot links \cite{Mehta_2008}. In this paper we model the locomotion with the assumption of anisotropic friction. Using the maximum dissipation principle \cite{stewart2000rigid}, we have derived a more accurate anisotropic Coulomb friction model and apply this to simulate surface friction. For underwater robots, dynamics has been studied in the fluids of different Reynolds numbers \cite{Hatton2013, Pettersen_2017_review, Kelasidi_2014, Khalil_2007}. At low Reynolds numbers, viscous friction is the dominant reaction force. At high Reynolds numbers, added mass and drag (fluid dynamics) dominate. We add these reaction forces to our dynamics model to simulate different environments. 

A common method to evaluate gait efficiency is the Pareto curve which shows the trade-off between locomotion speed and energy consumption \cite{Mehta_2008, Kelasidi_2018, snake_gaits_ppo}. We use this method to create a baseline to evaluate our auto-generated gaits. 

\section{Dynamic Model}\label{sec:dynamics}

The dynamics of snake locomotion are well studied in the literature, as a dynamic model is often used to validate or tune a proposed gait. In our work, the dynamic model is also used to generate the trajectory, via optimization with respect to a given cost function. Before describing the gait synthesis method, we will thus describe the dynamic model we use, which follows a standard time-stepping method based on a Newton-Euler formulation with explicit integration for a 2D n-link kinematic chain.

\begin{figure}[t]
\centerline{\includegraphics[clip, width=0.8\linewidth]{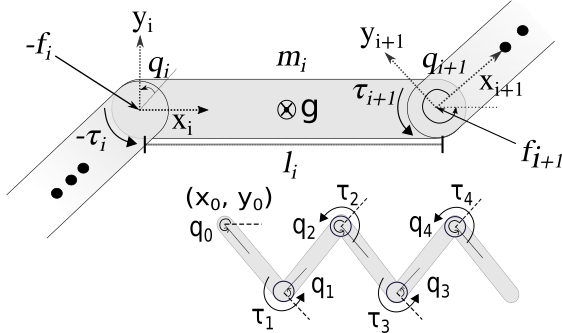}}
\caption{Illustration of the structure and dynamic model of the snake robot. The top figure shows link $i$, its axes, torques, mass, and length. The bottom figure shows an example of a 5-link snake.}
\label{link_diagram}
\end{figure}

We employ the model depicted in Fig.~\ref{link_diagram}, where $q_0$ is the angle of the first link with respect to the world, and $q_i$ denotes the relative angle of link $i$ with respect to the previous link (counterclockwise); $\tau_i$ is the motor torque applied to joint $i$; and $\mathbf{p}_0 = [x_0, y_0]$ is the head position in the global coordinates. The length of link $i$ is $l_i$ and its mass is denoted by $m_i$. We assume uniform distribution of mass so the center of gravity is located in the middle of each link; gravity is pointing into the page. Each link has a local coordinate frame based at the proximal end, with the $y_i$ axis pointing in the longitudinal direction and the $x_i$ axis in the transverse direction. 

According to Newtonian dynamics, at any time step, force equilibrium on link $i$ expressed in the local coordinate frame requires:
\begin{eqnarray}
-\mathbf{f}_i + R_i^{i+1} \mathbf{f}_{i+1} &=& m_i \mathbf{a_{ci}} + \mathbf{f}^{env}_i \label{eq:force} \\
\mathbf{a_{ci}} &=& \mathbf{a_{i}} + 0.5 l_i \dot\omega_{i}\begin{bmatrix} 0 \\ 1 \end{bmatrix} + 0.5l_i \omega_i^2 \begin{bmatrix} -1 \\ 0 \end{bmatrix}
\end{eqnarray}
where $\mathbf{f}_i$ is the internal force applied by link $i+1$ to link $i$ in the local coordinates; $R_i^{i+1}$ is the rotation matrix from the local coordinates $i+1$ to the local coordinates $i$; $\mathbf{a_{ci}}$ is the acceleration of the center of mass of link $i$; $\mathbf{a_i}$ is the linear acceleration of joint $i$; $\mathbf{\omega_i}$ is the angular velocity of joint $i$; $\mathbf{\dot\omega_{i}}$ is the angular acceleration of joint $i$. Of particular interest is the the vector $\mathbf{f}^{env}_i$, which denotes external forces applied by the environment in response to the snake movement. We dub these "reaction forces", and discuss them in detail in the following section.

Similarly, torque equilibrium on link $i$ requires:
\begin{equation}
-\tau_i + 0.5l_i\begin{bmatrix}0&1\end{bmatrix}\left(\mathbf{f}_i + R_i^{i+1}\mathbf{f}_{i+1}\right) + \tau_{i+1} = \dot{q}_i \mu_v 
\end{equation}
where $\tau_i$ is the torque applied by link $i+1$ to link $i$; $\mu_v$ is the joint viscous friction coefficient. Additional constraints follow from the consistency of angular and linear accelerations between consecutive links, with no grounding on the first joint:
\begin{eqnarray}
\dot\omega_{i-1} + \ddot{q}_i - \dot\omega_i &=& 0 \\
R_i^{i+1}(\mathbf{a_{i-1}} + l_{i-1}\dot\omega_{i-1}\begin{bmatrix} 0 \\ 1 \end{bmatrix} +  l_{i-1} \omega_{i-1}^2 \begin{bmatrix} -1 \\ 0 \end{bmatrix}) &=& \mathbf{a_{i}} \\ 
\mathbf{f}_0 &=& 0 \label{eq:ground}
\end{eqnarray} 

For known motor torques $\tau_i$, we solve the system (\ref{eq:force})-(\ref{eq:ground}) for unknowns $\mathbf{f}_i,~\dot{\omega}_i,~\mathbf{a}_i,~\ddot{q}_i$. We use numerical integration to advance to the next time step: computing ${q}_i,~\dot{q}_i$ from $\ddot{q}_i$ and $\mathbf{p}_0$ from $\mathbf{a}_0$. Note that multiple numerical integration methods are used in our work; we use the more accurate RK4 method when using this model for validation, and the faster Euler method when using the model for gait synthesis.
\section{Modeling Environments}\label{sec:reaction_forces}

Biological snakes can achieve undulatory motion in many types of environments (e.g. overland, underwater) and at many scales. In each case, they achieve this by creating or leveraging anisotropic effects in the reaction forces applied by the environment in response to their motion. However, the exact nature of these forces varies greatly between environments, giving rise to different strategies: during locomotion, some snakes adjust their scales, other redistribute their weight or change their winding angles, etc.~\cite{filippov2018numerical}. These strategies produce different gaits, each suited to its environment. 

We can attempt to capture the characteristics of a given environment by using an appropriate model for the reaction forces applied in our dynamics formulation described previously. We use the broader term "reaction forces" to encompass both frictional and non-frictional effects which contribute to locomotion. In this work, we use three such models, described below. 

\subsection{Dry friction}

The most commonly used model for simulating dry frictional contact is that of Coulomb friction~\cite{STEWART96,stewart2000rigid}. For undulatory motion over solid, dry surfaces (e.g. overland snakes not moving on granular media), anisotropic effects can be captured by using different friction coefficients for longitudinal vs. transverse friction forces. However, the Coulomb friction model introduces non-linear and non-convex constraints that are difficult to solve efficiently. A simplified version, similar to the model used in previous studies on snake robot locomotion \cite{saito2002, liljeback2012snake, Hirose_1993} is: 
\begin{equation}
\mathbf{f}^{env}_{box} = -mg
\begin{bmatrix} 
\mu_l & 0 \\
0 & \mu_t 
\end{bmatrix}
sgn(\begin{bmatrix} 
v_l \\
v_t
\end{bmatrix})
\label{matei_friction}
\end{equation}
where $\mu_{t}$ is the transversal and $\mu_{l}$ is the longitudinal coefficients of friction. And $v_t$ is the transversal and $v_l$ is the longitudinal component of the velocity of the center of mass. 

However, this simplified model has multiple limitations. First, applied friction direction as a function of contact velocity (relative velocity of the link w.r.t the ground) is non-smooth. Second, it does not obey the Maximum Dissipation Principle~\cite{stewart2000rigid}, which states that, given a contact velocity $v$, the ensuing friction force $\mathbf{f}$ maximizes energy dissipation (i.e., $-\mathbf{f}v$), subject to other constraints (such as friction coefficients). We thus introduce a new smooth model for anisotropic dry friction that obeys the Maximum Dissipation Principal, formulated as:
\begin{equation}
\mathbf{f}^{env}_{dry} = -mg
\begin{bmatrix} 
\mu_l & 0 \\
0 & \mu_t 
\end{bmatrix}
\begin{bmatrix} 
sin(arctan(\frac{\mu_t v_t}{\mu_l v_l})) \\
cos(arctan(\frac{\mu_t v_t}{\mu_l v_l}))
\end{bmatrix}
\label{max_friction}
\end{equation}

\begin{figure}
\centerline{\includegraphics[width=0.6\linewidth]{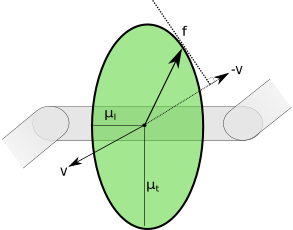}}
\caption{Proposed dry friction model. Selected friction direction maximizes energy dissipation under anisotropic friction constraints.}
\label{coulomb_diagram}
\vspace{-4mm}
\end{figure}

Eq.~\eqref{max_friction} is derived by maximizing the dissipation $-\mathbf{f}v$ with the assumption that the anisotropic friction is represented by a friction ellipse with the major axis equal to $\mu_{t}$ and the minor axis equal to $\mu_l$ as shown in Fig.~\ref{coulomb_diagram}.

\subsection{Viscous friction}
For underwater undulatory locomotion, anisotropic viscous friction plays an important role for generating forward motion. At low Reynolds number (i.e. small scale organisms in water), it can be considered the dominant reaction force. A model of reaction forces that comprises viscous friction exclusively can be considered as an approximation for small scale swimmers, such as tadpoles. We note that viscous friction has occasionally been used in the literature to model overland snakes as well~\cite{Pettersen_2017_review}, even though it is a less accurate approximation of dry Coulomb friction than the box model discussed above. We model viscous friction as:
\begin{equation}
\mathbf{f}^{env}_{viscous}= -
\begin{bmatrix} 
c_l & 0 \\
0 & c_t 
\end{bmatrix}
\begin{bmatrix} 
v_l \\
v_t
\end{bmatrix}
\label{anisotropic_viscous}
\end{equation}
where $c_l$ and $c_t$ are the longitudinal viscous friction coefficient and the transversal friction coefficient respectively.
\subsection{Added mass}
In fluids of high Reynolds number, drag and added mass effect are the dominating reaction force. Added mass is particularly important for modeling large undewater snakes or eels because their bodies displace a significant volume of water during locomotion. 

We assume the links are cylindrical with a cross section of height $a$ and width $b$. The length of the link is denoted by $l$. We compute the drag force  \cite{Kelasidi_2014} \cite{Khalil_2007} as:
\begin{equation}
\mathbf{f}^{env}_{fluids} = -
\begin{bmatrix} 
\frac{1}{2}pC_dal & 0 \\
0 & \frac{1}{2}p\pi C_f\frac{a+b}{4}l
\end{bmatrix}
\begin{bmatrix} 
sgn(v_l)v_l^2 \\
sgn(v_t)v_t^2
\end{bmatrix}
\label{drag}
\end{equation}
where $p$ is fluid density, and $C_f$ and $C_d$ are the drag coefficients in the longtitudinal and transverse directions. 

In addition, the volume of fluid displaced during motion has the effect equivalent to adding mass to the links of the snake. Due to the shape of the link, this effect is anisotropic, with more mass added for motion in the transverse direction. Therefore, in  Eq.~\eqref{eq:force}, we must distinguish between added mass in the tranverse vs. longitudinal directions:
\begin{eqnarray}
-\mathbf{f}_i + R_i^{i+1} \mathbf{f}_{i+1} &=& 
\mathbf{M}_i\mathbf{a_{ci}} - g\mathbf{M}_iR_w^i\begin{bmatrix} 0 \\ -1 \end{bmatrix} + \mathbf{f}^{env}_i 
\end{eqnarray}
where we compute the added mass using:
\begin{eqnarray}
\mathbf{M}_i &=&
\begin{bmatrix} 
m_i & 0 \\
0 & m_i + m_i^{add} 
\end{bmatrix}
\\
m_i^{add} &=& p\pi C_a\frac{a^2}{4}l
\label{added_mass}
\end{eqnarray}

where $C_a$ is the added mass coefficient.

\section{Gait Synthesis Method}\label{sec:methods} 

For gait synthesis, we use a Model Predictive Control (MPC) method which relies on an internal model of the dynamics of the system in order to repeatedly optimize a control sequence. A full review MPC is beyond the scope of this paper; we provide here a short overview, then focus on applicability to snake locomotion. Given a (potentially nonlinear) dynamic model 
\begin{equation}\label{nonlinear_sys}
\mathbf{x}(t+1) = f(\mathbf{x}(t), \mathbf{u}(t))
\end{equation}
where $\mathbf{x}$ denotes the state and $\mathbf{u}$ denotes the control action, MPC computes the optimal action sequence $\mathbf{u}(i)$ ($i=t,\ t+1,\cdots, t+N-1$) that minimizes a cost function 
\begin{equation}\label{mpc_cost}
J = \sum_{i=t}^{t+N-1}l(\mathbf{x}(i), \mathbf{u}(i)) + l_f(\mathbf{x}({t+N}))
\end{equation}
where $N$ is the horizon length; $l(\mathbf{x}(i),\mathbf{u}(i))$ is the cost of the state and action at time $i$, and $l_f(\mathbf{x}(t+N)$ is the cost on the final state of this horizon. The specific optimization algorithm we use is iLQR~\cite{tassa2012iLQR}, a variant of the iterative Linear Quadratic Regulator previously demonstrated on problems such as legged locomotion and manipulation. 

After an optimized trajectory has been computed, the robot takes the first (or the first few) action(s) therein, and the process restarts from the new robot state. This allows the robot to compensate for errors which inevitably arise when following a given trajectory, due to the linearization of the dynamics model, or simply due to errors in the model.

\subsection{Application to Gait Synthesis}

For an n-link snake robot, we define the state vector as $\mathbf{x}=\begin{bmatrix} x_0 & y_0 & q_0 & \cdots & q_l & \dot x_0 & \dot y_0 & \dot q_0 & \cdots & \dot q_n \end{bmatrix}^T$, and the action vector $\mathbf{u} = \begin{bmatrix} \tau_1 & \cdots & \tau_n \end{bmatrix}$. The dynamic model $f(\mathbf{x}(t), \mathbf{u}(t))$ is computed as described in Sec.~\ref{sec:dynamics}. We use finite differencing to compute the derivatives of this model w.r.t. the state and action vectors, both of which are needed by the optimization routine.

For generating locomotion gaits without obstcles, we use the following quadratic cost functions:
\begin{eqnarray}\label{quadratic_cost}
l(\mathbf{x}, \mathbf{u}) &=& \alpha \|\mathbf{p}_{goal}-\mathbf{p}_0\| + \beta \sum_{j=1}^{n}\tau_{j}^2 \\
\label{quadratic_cost_final}
l_f(\mathbf{x}) &=& \alpha \|\mathbf{p}_{goal}-\mathbf{p}_0\|
\end{eqnarray}
where $\mathbf{p}_{goal}$ = $[x_g, y_g]$ is the desired position of the head of the snake, $\mathbf{p}_{0}$ = $[x_0, y_0]$ is the current position of the head of the snake, and $\alpha$ and $\beta$ are tunable parameters. 

For obstacle avoidance, we add an additional cost term that takes into account the relative position of each segment of the snake robot with respect to each obstacle. In other words, all of the obstacles in the environment and all of the segments of the snake contribute to this term. The distance between each obstacle and each snake segment is calculated as $d_{ij}$, where $i$ is the index for obstacles, and $j$ is the index for segments. We note that negative distance indicates that a snake segment is interpenetrating an obstacle. The obstacle cost is the sum of 2-dimensional Heaviside step functions for every $d_{ij}$:
\begin{equation}
l^{obs}(\mathbf{x}, \mathbf{u}) = \sum_{i=1}^{k}\sum_{j=1}^{n}{\frac{A}{1 + e^{2kd_{ij}}}} \label{eq:obs}
\end{equation}
where $k$ denotes the number of obstacles, and $A$ is a parameter that determines the rise of the step function. In all of our experiments we set $A = 1$.

\section{Results}\label{sec:experiments}

We test the proposed gait generation method using a five link snake robot in different environments, each based on one of the reaction forces models described in Sec.~\ref{sec:reaction_forces}. The parameters of the snake robot and reaction forces are summarized in Table~\ref{tab:dynamic_params}. Torques at the joints are capped at 1 Nm. All gaits are tested in simulation using the model described in Sec.~\ref{sec:dynamics} with RK4 integration.

\begin{table}[t]
\setlength{\tabcolsep}{3pt}
\caption{Parameters of the Snake Robot and Reaction Forces}
\begin{center}
\begin{tabular}{|c|c|c|c|c|c|c|c|c|c|c|c|}
\hline
$l_i$ [m] & $a$ [m]& $b$ [kg] & $m_i$ & $\mu_t$ & $\mu_l$ & $c_t$ & $c_l$ &$C_f$ & $C_d$ & $C_a$ & $p$ [kg$\cdot m^3$]\\
\hline
.2 & .15 & .05& .2 & .9 & .1 & 1 &  10 & .01 & 1 & 1 &  1000  \\
\hline
\end{tabular}
\label{tab:dynamic_params}
\end{center}
\vspace{-7mm}
\end{table}

\subsection{Baseline: Serpenoid Curve Gaits}

As a comparison baseline for the generated gaits, we will compare performance to that obtained by using pre-defined serpenoid curves. For each joint $i$, desired joint angle as a function of time is defined as:
\begin{equation}\label{serpenoid_curve}
q_i(t)=\alpha sin(2\pi ft + (i-1)\beta) + \gamma 
\end{equation}
where $f$, $\alpha$, and $\beta$ are the parameters that can be used to tune the behavior of the resulting gait. In practice, this desired trajectory at each joint can be tracked using PID control, resulting in locomoting gaits.

One of the key difficulties of using serpenoid curves is the need to tune the curve parameters to obtain the desired behavior. To find the most effective gaits, we carried out a grid search over curve parameters; the boundaries and step size of the grid search are listed in Table \ref{grid_search_params}. We found the performance of the resulting gaits to also be highly sensitive to the gains of the controller used to track the serpenoid curve; we thus included these gaits in our grid search.

\begin{figure*}[t]
	\centerline{\includegraphics[width=0.9\textwidth]{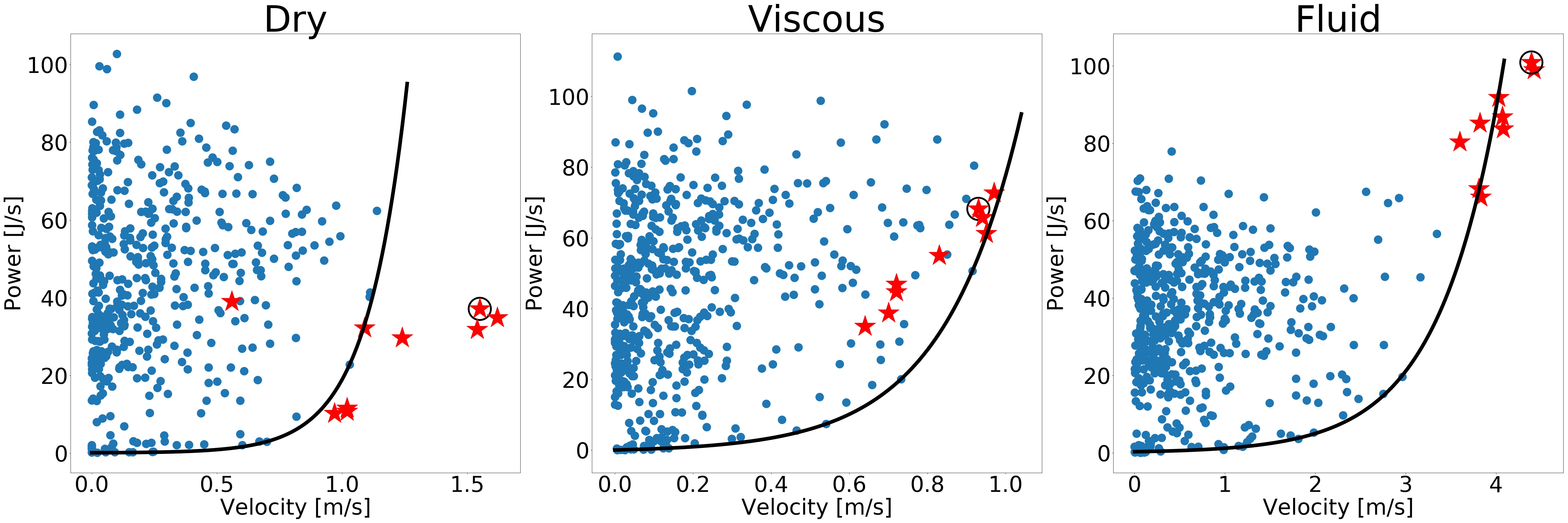}}
	\caption{Distribution of power consumption vs the forward speed. Each blue dot represents a serpenoid curve gait with a unique set of parameters $\alpha$, $\beta$, and $f$ where the PD controller was sucessfully tuned by the grid search (grid search values in Table. \ref{grid_search_params}.  The black line represents the Pareto-optimal front for the serpenoid curve gaits. The red stars represent MPC generated gaits. The stars circled in black represent a single set of gaits produced with the same set of MPC parameters: $\alpha$ = 1.0, $\beta$ = 0.1, and $x_g$ = -10.}
\label{velocity_vs_power_fig}
\vspace{-0mm}
\end{figure*}

We evaluated all gaits in our grid search in terms of speed and power consumption. For each combination of parameters, we generated a 6 s trajectory, with the snake starting from rest. To minimize the effect of the chosen start pose on gait performance, we considered the first 2 s of the trajectory to be "ramp up" time, and only measured speed and power over the last 4 s. Fig.~\ref{velocity_vs_power_fig} plots the power consumption versus the forward speed of each serpenoid curve (blue circles). The black line is the Pareto-optimal front \cite{belegundu2019optimization} that represents the best tradeoff between power consumption and locomotion speed. Gaits on this Pareto curve produce the corresponding forward speeds with the lowest power consumption. 

\begin{table}[t]
\caption{Serpenoid Curve and PD Controller Grid Search Parameters}
\begin{center}
\setlength{\tabcolsep}{4pt}
\begin{tabular}{cc}
\begin{tabular}{|c|c|c|c|}
\hline
\textbf{Gait}&\multicolumn{3}{|c|}{\textbf{Grid Search Space}} \\
\cline{2-4} 
\textbf{Param} & \textbf{\textit{Min}}& \textbf{\textit{Max}} & \textbf{\textit{Interval}} \\
\hline
$f$      & 0.5      & 10.0     & 0.25     \\
$\alpha$ & 0.1$\pi$ & 1.0$\pi$ & 0.1$\pi$ \\
$\beta$  & 0.5$\pi$ & 4.0$\pi$ & 0.1$\pi$ \\
\hline
\end{tabular}
&
\begin{tabular}{|c|c|c|c|}
\hline
\textbf{}&\multicolumn{3}{|c|}{\textbf{Grid Search Space}} \\
\cline{2-4} 
\textbf{Gain} & \textbf{\textit{Min}}& \textbf{\textit{Max}} & \textbf{\textit{Interval}} \\
\hline
$P$ & 0.1  & 3.0 & 0.1  \\
$D$ & 0.05 & 0.2 & 0.01 \\
\hline
\end{tabular}
\end{tabular}
\label{grid_search_params}
\end{center}
\vspace{-7mm}
\end{table}

\subsection{Synthesized Gaits and Environment Generalizability}

We used the proposed MPC-based method to synthesize gaits accross all three environments. We tested 9 sets of parameters of the quadratic cost Eq.~\eqref{quadratic_cost}~\eqref{quadratic_cost_final}, i.e., $\alpha=1$, $\beta=0.1, \ 0.01, \ 0.001$, $x_g = -10, \ -20, \ -50$ and $y_g =0$. The MPC horizon length N = 25. The speed and power of the resulting gaits are plotted in Figure ~\ref{velocity_vs_power_fig} as red stars. The gaits with $\alpha=1$, $\beta=0.01$, and $x_g = -20$ are circled in black, to illustrate the performance of a unique set of parameters accross all three environments. 

These results confirm MPC can synthesize effective locomotion gaits in all tested conditions. Furthermore, the resulting gaits are comparable to, and occasionally better than, their most efficient serpenoid curve counterparts. For example, in the dry friction environment, 3 sets of the MPC implementation generate gaits with speed near 1.6 m/s and power near 35 W. The Pareto-optimal gaits with a similar level of power can only achieve a speed of 1.1 m/s. In the viscous fricion and the fluid dynamics environments, all MPC-generated gaits are along the Pareto optimal trade-off between the power consumption and the speed. 

Beyond efficiency, we believe that one of the most attractive characteristics of automated gait synthesis is its generalizability accross different environments. We note that a single set of parameters ($\alpha=1$, $\beta=0.01$, and $x_g = -20$, circled in black in Fig. \ref{velocity_vs_power_fig}) produces effective gaits for all environments. In contrast, comparable behavior can be obtained with serpenoid curves only with environment-specific parameter tuning.

To obtain further insight, we studied the gaits produced by MPC across each environment with a single set of parameters ($\alpha = 1$, $\beta =0.01$) by decomposing the motion of each joint using a Discrete Fourier Transform (DFT); we computed the dominant frequency component $f_i$ and the corresponding amplitude $\alpha_i$ for each joint $i$ in the respective gait. The results are listed as in Table~\ref{mpc_frequency}. In each environment, the dominant joint frequency remains constant; this resembles the serpenoid curve in Eq. ~\eqref{serpenoid_curve}. However, the gaits differ from the serpenoid curve in amplitude variation. For the gait produced using fluid dynamic effects, the joints at the tail of the snake move at a significantly higher amplitude that the ones at the head, much more so than in the other two environments. This qualitatively mirrors behaviors observed in nature for similar enviroments, as in the case of eel-like anguilliform gaits~\cite{mclssac2003} and fish-like carangiform gaits~\cite{morgansen2001}. 

\begin{table}[t]
\setlength{\tabcolsep}{4pt}
\caption{DFT of MPC Gaits in Different Environments}
\label{mpc_frequency}
\begin{center}
    \begin{tabular}{ | c | c | c | c | c | c | c | c | c |}
    \hline
    \textbf{}&\multicolumn{4}{|c|}{\textbf{Frequency(Hz)}}&\multicolumn{4}{|c|}{\textbf{Amplitude(rad)}} \\
    \cline{2-9}
    \textbf{Environment} & $f_1$ & $f_2$ & $f_3$ & $f_4$ & $\alpha_1$ & $\alpha_2$ & $\alpha_3$ & $\alpha_4$ \\ \hline
    $Dry$ & 6.31 & 6.31 & 6.31 & 6.31 & 0.25 & 0.32 & 0.37 & 0.42\\ \hline
    $Viscous$ & 6.75 & 6.75 & 6.75 & 6.75 & 0.53 & 0.62 & 0.67 & 0.83 \\ \hline
    $Fluid$ & 5.25 & 5.25 & 5.25 & 5.25 & 0.51 & 0.76 & 1.40 & 2.42 \\ \hline
    \end{tabular}
\end{center}
\vspace{-7mm}
\end{table}

\subsection{Automatic Generation of Complex Gaits}

The gaits we have shown so far to illustrate our method, while efficient and generalizable accross different environments, are still serpenoid-like in nature. However, the proposed method for gait synthesis can also generate more complex gaits, without having to adhere to the serpenoid template. This can be achieved with the use of different cost functions or goal settings. 

\begin{figure}[t]
\vspace{-6mm}
\begin{tabular}{c}
	\centerline{\includegraphics[width=\linewidth]{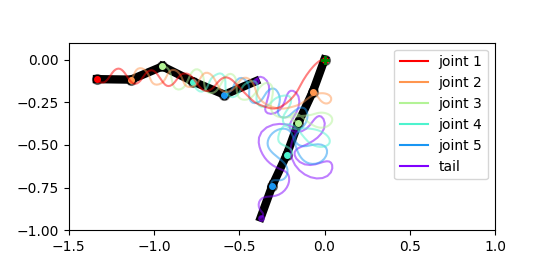}}\\[-2mm]
	{\footnotesize Sharp Turning Gait}\\[-0mm]
	\centerline{\includegraphics[width=\linewidth]{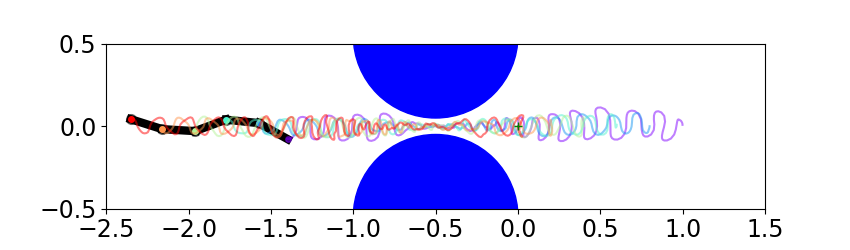}}\\[-2mm]
	{\footnotesize Obstacle Avoidance Gait Through a Narrow Tunnel}
\end{tabular}
\caption{MPC generates flexible gaits adaptive to environments. The top figure demonstrates a sharp turning gait. The bottom figure demonstrates obstacle avoidance through a narrow tunnel.}
\label{fig:complex}
\vspace{-5mm}
\end{figure}

Figure~\ref{fig:complex} illustrates this concept. The top image shows a gait generated using Eq.~\eqref{quadratic_cost}~\eqref{quadratic_cost_final}, but with a goal position for the snake head that induces a sharp turn (as opposed to a goal set in front of the snake, as before). The result is an effective turning motion that does not rely on a single-frequency serpenoid pattern. The bottom image illustrates the use of the obstacle-based component of the cost function shown in Eq.~\eqref{eq:obs}. Here, the gait is still serpenoid in nature, but the amplitudes of the different joints automatically adjust as the snake moves through the tunnel. At any point, the links outside the narrowest point of the tunnel are used to propel forward, while the links in the tunnel are kept still to avoid collision. Both gaits are also shown in the video accompanying the submission, as are additional results illustrative of our method.

\subsection{Practical Applicability}\label{sec:computation}

The proposed MPC-based gait generation method faces two important challenges for application to real robots: a reliance on dynamic models, and the computational cost. For real-life execution, the dynamic model will only be an approximation of real-life physics. Furthermore, gait generation must keep up with execution in order to be used online.

To study computational performance, we measured the time needed to optimize a single trajectory over a horizon of 25 steps. We use a control rate of 100 Hz: a 25-step trajectory represents 0.25 s of real life execution. We averaged computational performance over a complete 6 s locomotion gait for each environment. For dry friction, viscous friction and fluid dynamics, average computation times were 1.4 $\sec$s, 1.0 s, and 1.0 s respectively. Standard deviations where 1.4 s, 0.6 s, and 0.5 s respectively. All tests were performed on a commodity 16-core AMD Ryzen 7 PRO 1700 processor.

We note that our current implementation of iLQR is not efficient enough for real-life execution. For example, in the case of viscous friction, it takes 1 s of computation time to optimize a trajectory that takes 0.25 s to execute. We estimate, for practical applicability, a computation speed-up of at least 10X to 20X will be necessary; this may be achieved via an optimized implementation, more extensive parallelization, or algorithmic improvements. We will pursue all these directions in future work.

To study MPC's robustness to errors in the dynamics model, we performed a preliminary test by introducing errors into the coefficients of viscous friction (Eq. \ref{anisotropic_viscous}): we add a discrepancy between $c_t$ used by MPC to generate the gait, and $c_t$ used to test it. For a 5\% error in $c_t$, we observed a 2\% reduction in the trajectory speed; a 25\% error produced a 4\% slowdown. Becasue a dynamics model has numerous performance affecting parameters, a more complete evaluation of these effects is needed for real-life applications.

\section{Conclusions and Future Work}\label{sec:conclusions}

In this paper, we proposed a method of automatic gait synthesis for snake robots using iLQR, a trajectory optimization method from the MPC family. Unlike methods that use a serpenoid curve or similar patterns, this approach produces gaits without relying on predefined patterns, by optimizing the trajectory w.r.t. a given cost function. The key is the use of a model of the dynamics of the system, for which we use a standard time-stepping scheme with explicit integration and Newton-Euler chain dynamics at each time step. 

To encapsulate the characteristics of each environment, we used models of the anisotropic reaction forces that arise in response to snake movements. For the case of dry friction, we introduced a novel model that approximates anisotropic Coulomb friction with a smooth friction force vs. velocity profile, by integrating the Maximum Dissipation Principle. Other reaction forces models we used include viscous friction as well as fluid dynamic effects such as drag and added mass. Together, these reaction forces allow us to model three different environments for undulatory locomotion: overland on solid media, underwater in a low Reynolds number regime (small scale swimmers), and underwater in a large Reynolds number regime (large swimmers). It is, to the best of our knowledge, the first time that an automatic gait generation approach has been used to generate and analyze undulatory gaits across different environment models. 

Our key result shows that, without any tuning or change in parameters between environments, the proposed automatic gait synthesis method produces gaits that are as efficient as those obtained by manually tuning a serpenoid curve to each environment. The undulatory gaits produced by the proposed method also display some of the characterics observed in nature. However, as it does not depend on a pre-defined movement pattern, the same method can produce more complex gaits, e.g. for sharp turns or obstacle avoidance, by using different formulations for the optimized cost function.

Before these results can be used for gait generation on real robots, additional challenges still have to be tackled. In particular, a reduction in computation time is needed to meet the requirements of keeping up with real-life execution. Nevertheless, we believe that automatic gait generation methods as the one presented here can represent a useful complement to approaches based on a central pattern generator, as we aim towards robot snakes able to operate in complex, unstructured environments.





\bibliographystyle{IEEEtran}
\bibliography{bib/references}

\addtolength{\textheight}{-12cm}   

\end{document}